\documentclass[letterpaper, 10 pt, conference]{ieeeconf}  

\IEEEoverridecommandlockouts                              

\usepackage{cite}
\usepackage{makecell}                                    
\usepackage{booktabs}
\usepackage{microtype}
\usepackage{amsmath}

\usepackage{graphics} 
\usepackage{epsfig} 
\usepackage{mathptmx} 
\usepackage{times} 
\usepackage{amsmath} 
\usepackage{amssymb}  
\usepackage{caption}
\usepackage{multirow}

\title{\LARGE \bf
	Tactile Recognition of Both Shapes and Materials with Automatic Feature Optimization-Enabled Meta Learning
}

\author{Hongliang Zhao, Wenhui Yang, Yang Chen, Zhuorui Wang, Baiheng Liu and Longhui Qin%
	\thanks{This work was supported by the State Key Laboratory of Autonomous Intelligent Unmanned Systems (ZZKF2025-2-5) and the State Key Laboratory of Robotics and Systems (HIT) (SKLRS-2025-KF-11). (Corresponding author: Longhui Qin)}%
	\thanks{Hongliang Zhao, Wenhui Yang, Yang Chen, Zhuorui Wang, and Baiheng Liu are with the School of Mechanical Engineering, Southeast University, Nanjing 211189, China.
			E-mail: zhaohongliang@msn.com; wenhui-yang@outlook.com.}%
	\thanks{Longhui Qin is with the School of Mechanical Engineering, Southeast University, Nanjing, 211189, China, the State Key Laboratory of Autonomous Intelligent Unmanned Systems, Beijing Institute of Technology, Beijing, 10081, China and the State Key Laboratory of Robotics and Systems, Harbin Institute of Technology, Harbin, 150001, China. (e-mail: lhqin@seu.edu.cn).}}
	
\begin{document}	
	\maketitle
	\thispagestyle{empty}
	\pagestyle{empty}
	
	\begin{abstract}		
		Tactile perception is indispensable for robots to implement various manipulations dexterously, especially in contact-rich scenarios. However, alongside the development of deep learning techniques, it meanwhile suffers from training data scarcity and a time-consuming learning process in practical applications since the collection of a large amount of tactile data is costly and sometimes even impossible. Hence, we propose an automatic feature optimization-enabled prototypical network to realize meta-learning, i.e., AFOP-ML framework. As a ``learn to learn" network, it not only adapts to new unseen classes rapidly with few-shot, but also learns how to determine the optimal feature space automatically. Based on the four-channel signals acquired from a tactile finger, both shapes and materials are recognized. On a 36-category benchmark, it outperforms several existing approaches by attaining an accuracy of 96.08\% in 5-way-1-shot scenario, where only 1 example is available for training. It still remains 88.7\% in the extreme 36-way-1-shot case. The generalization ability is further validated through three groups of experiment involving unseen shapes, materials and force/speed perturbations. More insights are additionally provided by this work for the interpretation of recognition tasks and improved design of tactile sensors. 
	\end{abstract}
	
	\section{INTRODUCTION}
	
	Tactile sensing is crucial for robotic dexterous manipulation in contact-rich scenarios, especially when the susceptible visual information is limited or inaccessible in case of dim light or object occlusion. The powerful capability of tactile perception has been demonstrated in a variety of applications, such as slip detection for stable grasps \cite{li2018slip}, measurement of both static and dynamic stimuli \cite{qin2023perception}, and interpretation of complex contact physics \cite{narang2021interpreting}. However, alongside the artificial intelligence-facilitated development of multifarious perception algorithms, it is suffering from how to acquire considerable amount of effective tactile data for model training, considering robotic experiments are costly while high-fidelity tactile simulation remains challenging \cite{el2024optimizing}.
	
	\begin{figure}[htbp!]
		\centering
		\includegraphics[width=1\columnwidth]{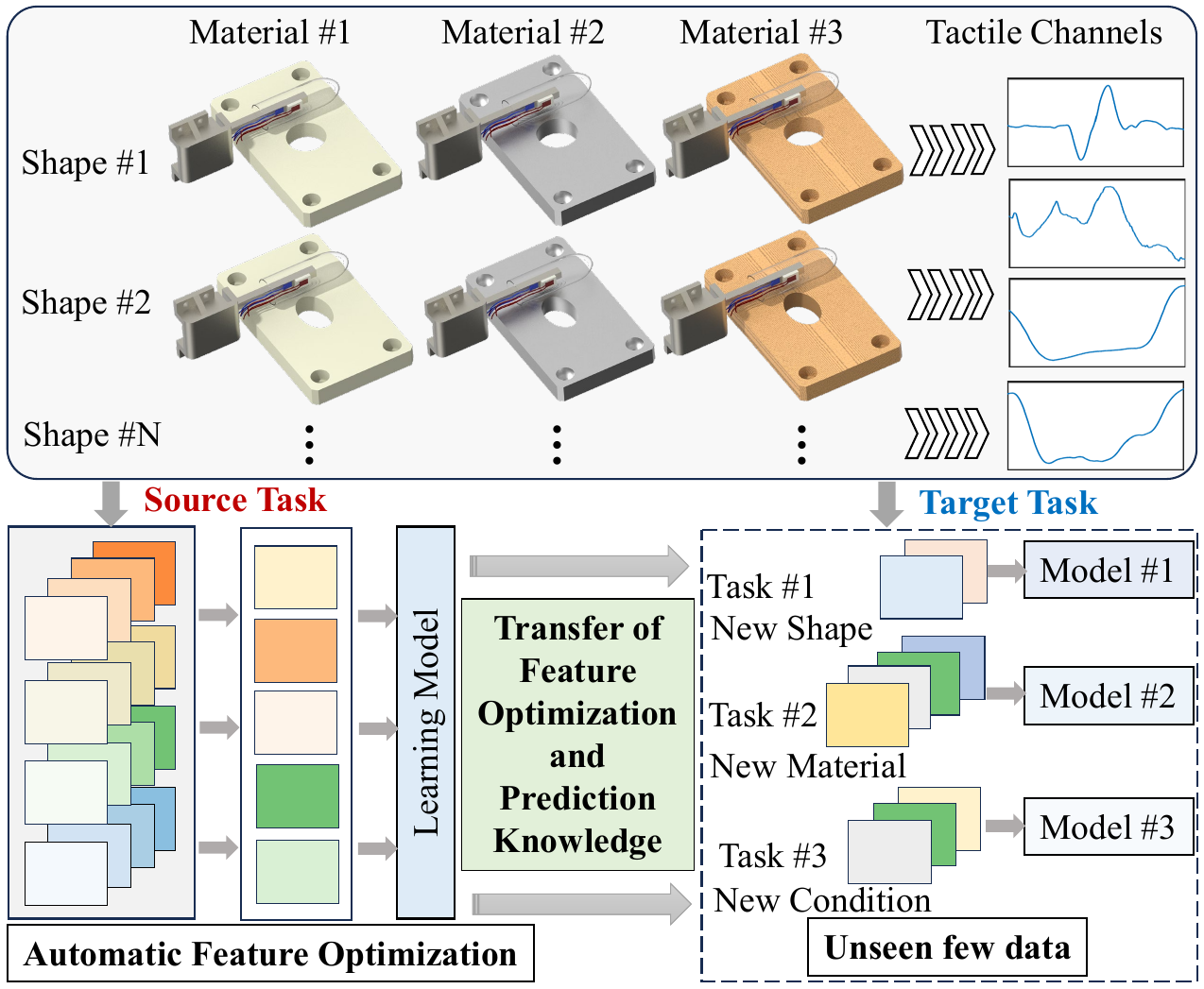}
		\caption{Conceptual overview of the proposed AFOP-ML framework. Multi-channel tactile signals produced by a tactile finger are firstly leveraged to learn how to optimize the feature space automatically and predict correct shape and material category as a source task. Then, these knowledge is transferred to the AFOP-ML model to recognize new shapes and materials under normal or changed experimental conditions as a target task with few-shot data available. Features that are fed into the prototypical network are automatically determined for different recognition tasks.}
		\label{fig:abstract}
	\end{figure}
	
	As data scarcity severely impacts the performance of traditional machine learning (ML) algorithms, e.g., leading to the overfitting or underfitting of ML models, deep learning (DL) has shown great success in numerous fields but adversely exhibits more data-hungry and computationally expensive disadvantages. Although transfer learning could relieve this predicament by fine-tuning a pretrained model on the given new task, it tends to result in poor generalization performance when the dataset size is very small or distinctive from the source tasks \cite{vettoruzzo2024advances}. In contrast, we human being are usually able to acquire new skills rapidly with few or even only one example. This phenomenon ``doing more with less" is benefited by our accumulated experience in the past, which taught us ``how to learn" to become a generalist. This kind of few-shot learning can be realized by meta-learning, one of the latest learning schemes in recent years, that learns to learn. In essence, it can be interpreted as a N-way-K-shot classification problem, where N-way refers to the N classes to discriminate and K-shot represents there are K samples in each class. Although its effects in tactile recognition has been demonstrated by very few pioneered works, the vital problem of feature extraction not only severely affects the accuracy, but impedes its wide application \cite{kaboli2016re, bauml2019deep}, since the hand-crafted method requires frequent human interference while automatic extraction by convolutional neural network (CNN) employs neither physically meaningful nor optimal features.
	
	In this paper, we propose an Automatic Feature Optimization-enabled Prototypical network to realize Meta Learning (AFOP-ML) in tactile recognition of both shapes and materials from minimal data, as illustrated in Fig.~\ref{fig:abstract}. Few-shot learning is successfully carried out relying on learning how to learn not only the prediction capability from constructed meta-training tasks, but also more importantly automatic determination of optimal feature combinations adaptively for each new task. Based on a comprehensive feature pool, Neighborhood Component Analysis (NCA) is leveraged to rank the importance of each physical features, and then an episodic scan is implemented to determine the optimal feature dimensionality, before feeding to a lightweight prototypical network \cite{snell2017prototypical} for rapid adaptation. When our bio-inspired tactile finger that comprises four sensing channels, two for static stimuli and the other two for dynamic, was applied to the recognition of both 12 shapes and 3 materials, AFOP-ML was found able to learn how to make choices automatically of the optimal features corresponding to different tasks. It shows distinctive behaviors in unseen shapes, materials and force/speed perturbations. In 5-way-1-shot case, it achieves an accuracy as high as 96.08\% while the consumption time is significantly reduced in pretraining and adaptation phases.
	
	The contributions are summarized as: At first, as far as we know, it's the first time that meta-learning is applied to tackle the tactile recognition of both shapes and materials using  a tactile finger working in piezoresistive and piezoelectric principles. Secondly, in addition to the learning of prediction models, the proposed framework also learns adaptive determination of the optimal feature space for different tasks. Thirdly, its effectiveness and generalization are successfully demonstrated by a series of tactile recognition tasks involving unseen shapes, materials and physical perturbations.
	
	\section{Related Works}
	
	\subsection{Tactile Recognition}
	Bio-inspired by the mechanoreceptors in human skin, plentiful of tactile sensors have emerged working in various principles, e.g., piezoelectric, piezoresistive, triboelectric, and visuotactile sensing manners etc \cite{pyo2021recent, roberge2023stereotac}. Therein, sensing element (SE), as the essential component, converts the physical quantity into measurable electronic signals or images. Artificial skin is characterized by its lightweight and sensitive advantages, and can be glued to robotic hand directly \cite{bauml2019deep}. Some commercial sensors, such as the BioTac sensor, can be mounted at the tips of robotic hands \cite{kaboli2016re}. Tactile fingers that resemble human finger in the appearance and structure, not only provide the functionality of tactile perception, but play the role of end-effectors \cite{qin2017enhanced}.
	
	Combined with different algorithms, plenty of tactile perception tasks can be implemented. Conventionally, abundant handcrafted, physics-informed features, such as statistical and spectral descriptors, are extracted to feed into shallow neural network (NN) or ML models \cite{shi2023surface,wang2022fabric,qin2023perception}. The primary advantage lies in their exceptional data efficiency and interpretability, while at the same time frequent human interference is required and different features need to be tailored in new tasks. 	
	In contrast, end-to-end DL models utilized deep architectures to learn powerful feature representations automatically \cite{bottcher2021object,baishya2016robust}. Alternatively, tactile signals were converted into time-frequency images via Continuous Wavelet Transform (CWT) and then feature extraction was implemented with pretrained convolutional neural network (CNN) \cite{qin2024surface}. However, DL techniques are data-hungry and usually pose high requirement for the computational resources. All these methods will suffer from severe degradation of classification performance when the size of training data becomes very small, for example, few samples per class or even only one sample.
	
	\subsection{Meta-Learning for Tactile Perception}
	Meta-learning, or ``learning to learn," offers a compelling solution to the data scarcity problem by enabling models to rapidly adapt to novel tasks from only a few examples. It has shown significant potential in robotic perception \cite{hospedales2020meta}. There are three typical approaches: metric-based, model-based and optimization-based meta-learning \cite{vettoruzzo2024advances}. Two main stages are usually involved: meta-training to acquire generalized knowledge to solve new tasks, and meta-testing to evaluate the model performance in unseen samples. As for each task, the dataset is divided into support set and query set for model training and assessment.
	
	In the task of texture discrimination, 16 features from two finger sensors were employed and 10 prior texture models were constructed based on Least Squared Support Vector Machine (LS-SVM) \cite{kaboli2016re}. Then the nearest model was selected, updated and re-weighted for new tasks. Convolutional Siamese neural network was leveraged in \cite{bauml2019deep} to realize material recognition by parameter updating between two shared twin CNNs and utilizing its powerful ability of feature extraction. Optimization-based meta-learning, such as Model-Agnostic Meta-Learning (MAML) \cite{finn2017model} and its variants \cite{nichol2018reptile}, learns the parameter initialization that is highly sensitive to new tasks, allowing for effective adaptation with a few gradient steps. It can be found that most existing implementations depend on DL as their feature backbone \cite{bauml2019deep}, and often lead to weak interpretation of the feature representations, and reduction of prediction accuracy due to non-optimal feature combinations and high computational latency. Hence, in addition to the exploration of tactile recognition of both shapes and materials with a tactile finger, we also try to address the problem of how to learn the automatic selection of optimal features in meta-learning.
	
	\section{Tactile Finger and Data Collection}
	
	\subsection{Tactile Finger}
\begin{figure}[htbp!]
		\centering
		\includegraphics[width=1\columnwidth]{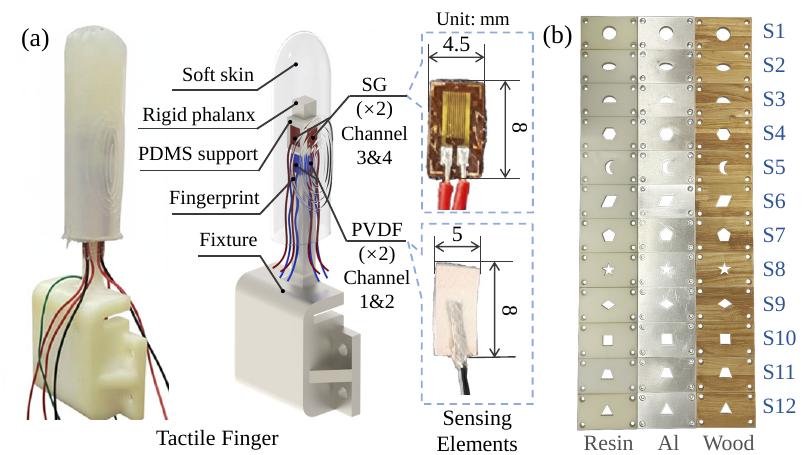}
		\caption{Tactile sensor and 36 categories to recognize. (a) Bio-inspired tactile finger mainly consists of rigid phalanx, soft skin, PDMS support, fingerprint, fixture and two types of SEs: PVDFs and SGs. Two PVDFs and two SGs constitute Channel 1 to 4. (b) The 36 categories to recognize. Three materials: Resin, Wood and Aluminum. There are 12 shapes for each material.}
		\label{fig:sensor_shape}
	\end{figure}

	A bio-inspired tactile finger is employed due to its soft-rigid-hybrid, low-cost and easy-to-fabricate characteristics. As illustrated in Fig. \ref{fig:sensor_shape}(a), it contains two types of SEs: two PVDFs (polyvinylidene fluorides, Tactile Channel 1\&2), and two SGs (strain gauges, Tactile Channel 3\&4). While Channel 1\&2 are responsible for dynamic stimuli, Channel 3\&4 correspond to static forces. Details about the finger can be found in \cite{zhao2025tactile}.	
	
	\begin{figure*}[h!]
		\centering
		\includegraphics[width=\textwidth]{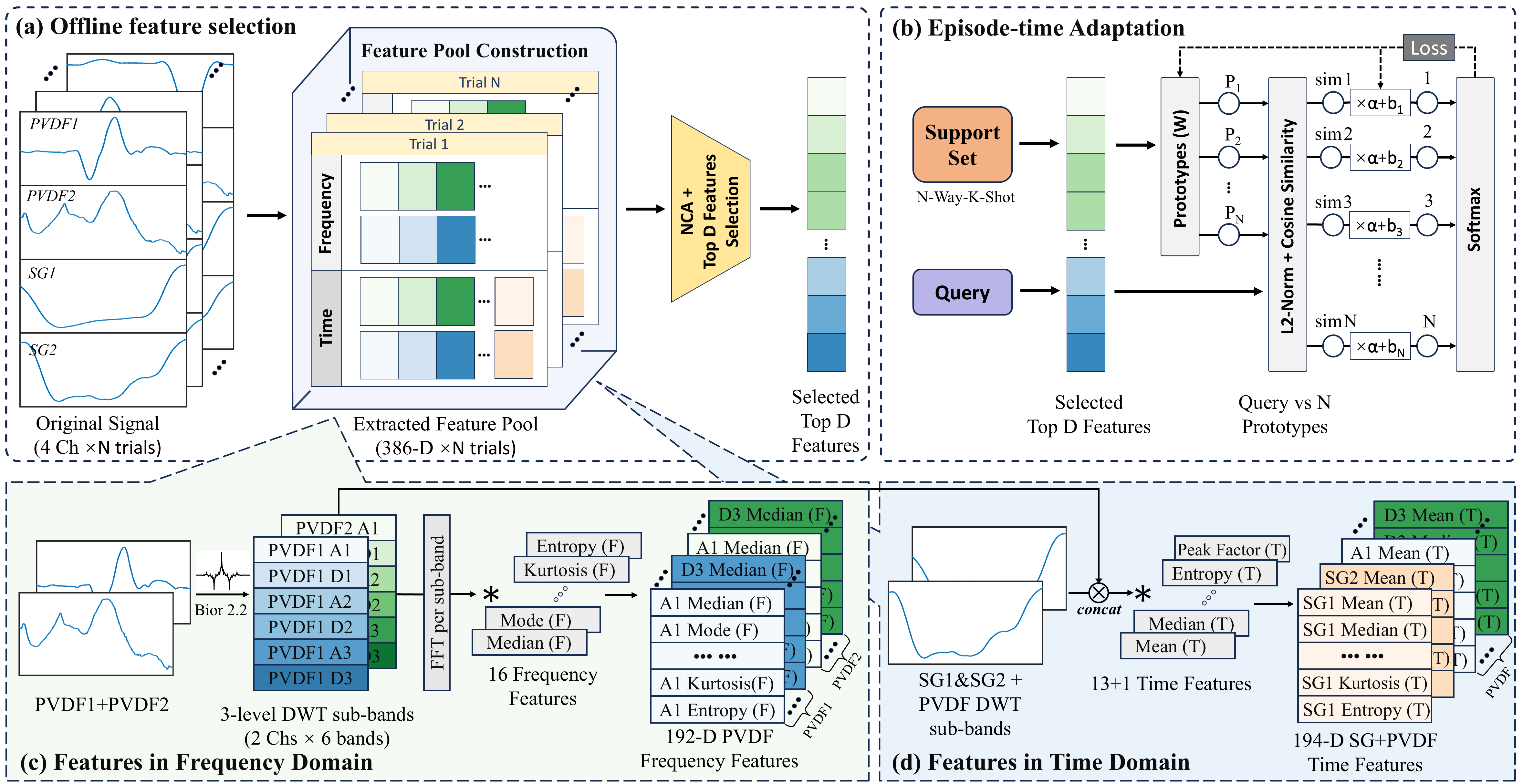}
		\caption{The framework of proposed AFOP-ML algorithm. (a) Offline feature selection. Four synchronous channels are converted to a 386-dimensionality feature pool per trial. (b) Episode-time adaptation. For each $N$-way-$K$-shot task, support sets are mapped to Top-$D$ and averaged into class prototypes $W$. Prototypes and queries are $\ell_2$-normalized; cosine similarities are temperature-scaled ($\alpha$) and shifted by class biases $b$, then fed to Softmax. Cross-entropy with entropy regularization updates only $(W,b)$ on the support set; queries are forward-only. (c) Features in frequency domain (192-D). (d) Features in time domain (194-D).}
		\label{fig:framework}
	\end{figure*}
	
	\subsection{Robotic Experiment and Tactile Data Collection}
	The tactile finger is mounted as an end-effector on the tip of a robotic arm (UR5, Universal Robot Inc.), which is controlled in ROS (Robot Operating System) platform. The 36 categories to recognize is shown in Fig. \ref{fig:sensor_shape}(b), including 3 materials (Resin, Wood, Aluminum), and 12 shapes (S1$\sim$S12: Circle, Ellipse, Semicircle, Hexagon, Moon, Parallelogram, Pentagon, Pentagram, Rhombus, Square, Trapezoid, Triangle) for each material.
	
	When the tactile finger was slid across each surface at 10 mm/s with constant contact force, the generated four-channel signals were recorded real-time by a data acquisition board (NI USB-6346, National Instruments) at 1 kHz. The sliding process for each category was repeated for 60 times. In subsequent analysis, time-series of data lasting for 2 seconds was employed, which was extracted from the contact phase.	
		
	\section{METHODOLOGY}
	
	\subsection{Meta-Learning Framework}
		
	The AFOP-ML framework is illustrated in Fig. \ref{fig:framework}, which is consisting of two main stages: an offline feature determination stage (Fig.~\ref{fig:framework}(a)) that identifies a compact and informative feature subset on the training dataset, and an episode-time adaptation stage (Fig.~\ref{fig:framework}(b)) that leverages the optimal features to implement rapid learning on new few-shot tasks.
	
	\subsection{Construction of Feature Pool}
	As detailed in Fig.~\ref{fig:framework}(c) and \ref{fig:framework}(d), we compute a 386-dimensional feature vector by combining 194 time-domain statistics and 192 frequency-domain metrics. The frequency features are derived from a 3-level Discrete Wavelet Transform (DWT) of the PVDF signals, capturing rich textural information across different frequency sub-bands. All features are then standardized using per-channel z-scoring. The specific definitions of these features stays the same with Ref. \cite{zhao2025tactile}.
	
	\subsection{Automatic Feature Optimization}
	Instead of selecting the most discriminative features manually, or determining the optimal dimension of feature space via time-consuming trial-and-error, we proposed to combine feature importance calculation and efficient dimension scan algorithm for automatic feature optimization. Here, ``automatic'' refers to data-driven ranking and selection within a pre-defined feature pool, rather than end-to-end feature learning from raw signals. Specifically, Neighborhood Component Analysis (NCA) was applied to produce the importance values for each feature, and then the optimal dimensionality of feature space, $D$, was determined with D-scan applied. As for each candidate $D$, 5-way-5-shot validation episodes were implemented on the training data using a prototypical classifier but without adaptation. The top-$D$ prefix of NCA ranking was progressively incremented to determine the optimal $D$ that maximizes recognition accuracy.
	
	\begin{figure}[hbpt!]
		\centering
		\includegraphics[width=0.9\columnwidth]{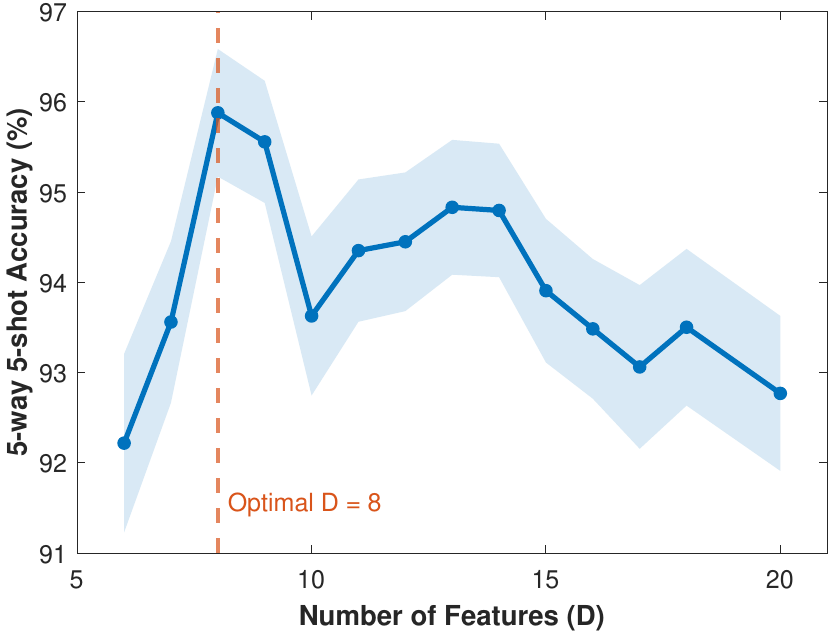}
		\caption{Automatic determination of the optimal dimensionality of feature space, $D$. 5‑way-5‑shot accuracy vs. the number of selected features (episodic mean ± 95\% confidence intervals (CI) across splits).}
		\label{fig:d_scan}
	\end{figure}
	
	The result of this scan for our 36-class closed-set task is presented in Fig.~\ref{fig:d_scan}, which indicates that insufficient features lead to an under-expressive model while excessive features may result in performance degradation due to overfitting on the small support sets. A clear peak emerges at $D=8$ revealing the optimal dimensionality as 8. It should be noted that the optimal dimensionality is not a fixed global hyperparameter but rather adapts based on the task's specific nature and difficulty. Detailed analysis of this adaptive behavior and its implications can be found in Sec.~\ref{subsec:interpretability}.
		
	\subsection{Episodic Classifier Backend}
	\label{sec:fep_backend}
	
	The lightweight prototypical network was employed to establish the AFOP-ML classifier backend, as shown in Fig.~\ref{fig:framework}(b). It adapts to new tasks in the metric-based approach and a cosine–softmax prototypical head is adopted with learnable parameter pairs.
	
	Given an episode containing a set of classes $\mathcal{C}=\{1,\dots,N\}$, the model is provided with a support set $S=\bigcup_{n\in\mathcal{C}} S_n$, where each $S_n$ contains $K$ training examples for class $n$. Each example is represented by $(\mathbf{x}_i, y_i)$, where $\mathbf{x}_i\in\mathbb{R}^D$ is a $D$-dimensional feature vector from above procedures and $y_i\in\mathcal{C}$ is its class label. The query set is denoted by $Q$. Bold lowercase letters denote vectors and bold uppercase letters denote matrices. For each class $n$, a prototype $\boldsymbol{\mu}_n$ is calculated as the mean of its support feature vectors:
	\begin{equation}
		\boldsymbol{\mu}_n \;=\; \frac{1}{|S_n|}\sum_{(\mathbf{x}_i,y_i)\in S_n} f_\phi(\mathbf{x}_i).
		\label{eq:prototype}
	\end{equation}
	where $f_\phi()$ denotes the mapping function from data to feature space. $\ell_2$-normalization is applied to both the query features and class prototypes. Let
	$\hat{\mathbf{x}}=\frac{f_\phi(\mathbf{x})}{\|f_\phi(\mathbf{x})\|_2}$ and
	$\hat{\mathbf{w}}_n=\frac{\boldsymbol{\mu}_n}{\|\boldsymbol{\mu}_n\|_2}$.
	The logit for class $n$ is
	\begin{equation}
		z_n(\mathbf{x}) \;=\; \alpha \,\langle \hat{\mathbf{x}}, \hat{\mathbf{w}}_n \rangle \;+\; b_n,
		\label{eq:cosine_logit}
	\end{equation}
	where $\alpha>0$ is a temperature (scale) and $b_n$ is a learnable bias.
	Class posteriors are given by
	\begin{equation}
		p(y=n \mid \mathbf{x}) \;=\; \frac{\exp(z_n(\mathbf{x}))}{\sum_{j=1}^{N}\exp(z_j(\mathbf{x}))},
	\end{equation}
	and queries are classified by
	\begin{equation}
		\hat{y} \;=\; \arg\max_{n}\, p(y=n \mid \mathbf{x}_q).
		\label{eq:protonet}
	\end{equation}

	At the beginning of each episode, the weight matrix $W\in\mathbb{R}^{N\times D}$ is
	row-wise initialized with the normalized prototypes $\hat{\mathbf{w}}_n^\top$ and the bias $b\in\mathbb{R}^{N}$ is set to zero.
	We then \emph{only} update $(W,b)$ on the support set (the feature extractor $f_\phi$ is frozen), by minimizing	
	\begin{equation}
		\mathcal{L}_{\text{support}}
		\;=\;
		\frac{1}{|S|}\sum_{(\mathbf{x},y)\in S}
		\Big[
		-\log p(y\mid \mathbf{x}) \;+\; \lambda\,\mathcal{H}\big(p(\cdot\mid \mathbf{x})\big)
		\Big],
	\end{equation}
	where $\mathcal{H}(p)=-\sum_j p_j\log p_j$ is the Shannon entropy and $\lambda\!\ge\!0$ is an entropy regularizer. Queries are evaluated with the adapted $(W,b)$; no gradients are taken on query samples. In this work, $\lambda=0.10$ (selected via a small sweep on held-out training episodes, where it achieved the best validation accuracy), and $(W,b)$ is optimized with the Adam (Adaptive Moment Estimation) optimizer for 250 steps at a learning rate of $1.5{\times}10^{-3}$.

		\begin{table*}[h!]
		\caption{CLOSED-SET ACCURACY AND EFFICIENCY COMPARISON}
		\label{tab:closed_set_results}
		\centering
		\resizebox{\textwidth}{!}{
			\begin{tabular}{l l ccccc ccc ccc cc}
				\toprule
				\multicolumn{2}{c}{} & \multicolumn{5}{c}{\textbf{1-Shot Acc (\%)}} & \multicolumn{3}{c}{\textbf{3-Shot Acc (\%)}} & \multicolumn{3}{c}{\textbf{5-Shot Acc (\%)}} & \multicolumn{2}{c}{\textbf{Time Cost (5-way-5-shot)}} \\
				\cmidrule(lr){3-7} \cmidrule(lr){8-10} \cmidrule(lr){11-13} \cmidrule(lr){14-15}
				\multicolumn{2}{c}{} &
				\textbf{5 Way} & \textbf{10 Way} & \textbf{12 Way} & \textbf{28 Way} & \textbf{36 Way} &
				\textbf{5 Way} & \textbf{12 Way} & \textbf{36 Way} &
				\textbf{5 Way} & \textbf{12 Way} & \textbf{36 Way} &
				\textbf{Pretrain} & \textbf{Adapt/episode (ms)} \\
				\midrule
				\multirow{5}{*}{\makecell[l]{\textbf{Meta-learning}\\\textbf{(episodic)}}}
				& \textbf{AFOP-ML (Ours)}         & \textbf{96.08} & \textbf{94.69} & \textbf{94.00} & \textbf{90.32} & \textbf{88.74} & 98.69 & 97.23 & \textbf{93.65} & 98.76 & 97.63 & 94.56 & \textbf{$\sim$2 s} & 391.04 \\
				& AFO-MLP-ML \cite{yoon2019tapnet}             & 93.47 & 90.09 & 87.82 & 75.79 & 69.64 & 97.99 & 95.56 & 85.78 & 98.82 & 96.25 & 86.22  & \textbf{$\sim$2 s} & 10241.72 \\
				& Direct-Prot-ML           & 92.05 & 88.79 & 86.18 & 80.92 & 78.47 & 96.82  & 93.56  & 87.89  & 97.22  & 95.20  & 89.87  & \textbf{$\sim$2 s} & 545.65 \\
				& MAML \cite{qiao2025fault}              & 93.28 & 86.70 & 84.38 & 73.99 & 70.77 & 94.88 & 88.61 & 63.89 & 95.37 & 89.32 & 68.06 & $\sim$20 min & \textbf{72.87} \\
				& CWT-ResNet-ML \cite{qin2024surface}           & 95.67 & 92.47 & 91.13 & 86.18 & 84.24 & \textbf{98.95} & \textbf{97.59} & 93.52 & \textbf{99.27} & \textbf{98.55} & \textbf{96.15}  & $\sim$8 min  & 1428.30 \\
				\addlinespace
				\multirow{2}{*}{\makecell[l]{\textbf{DL without} \\ \textbf{Meta-learning}}}
				& CNN               & 70.72 & 41.48 & 35.57 & 14.86 & 14.14 & 71.78 & 45.68 & 45.19 & 68.16 & 60.94 & 66.96 & N/A & 1643.25 \\
				& BiLSTM            & 68.85 & 44.02 & 39.20 & 16.78 & 21.71 & 73.59 & 55.63 & 32.85 & 70.43 & 59.60 & 44.26 & N/A & 14495.07 \\
				\bottomrule
		\end{tabular}}
	\end{table*}
	
	\begin{figure*}[hbt!]
		\centering
		\includegraphics[width=0.9\textwidth]{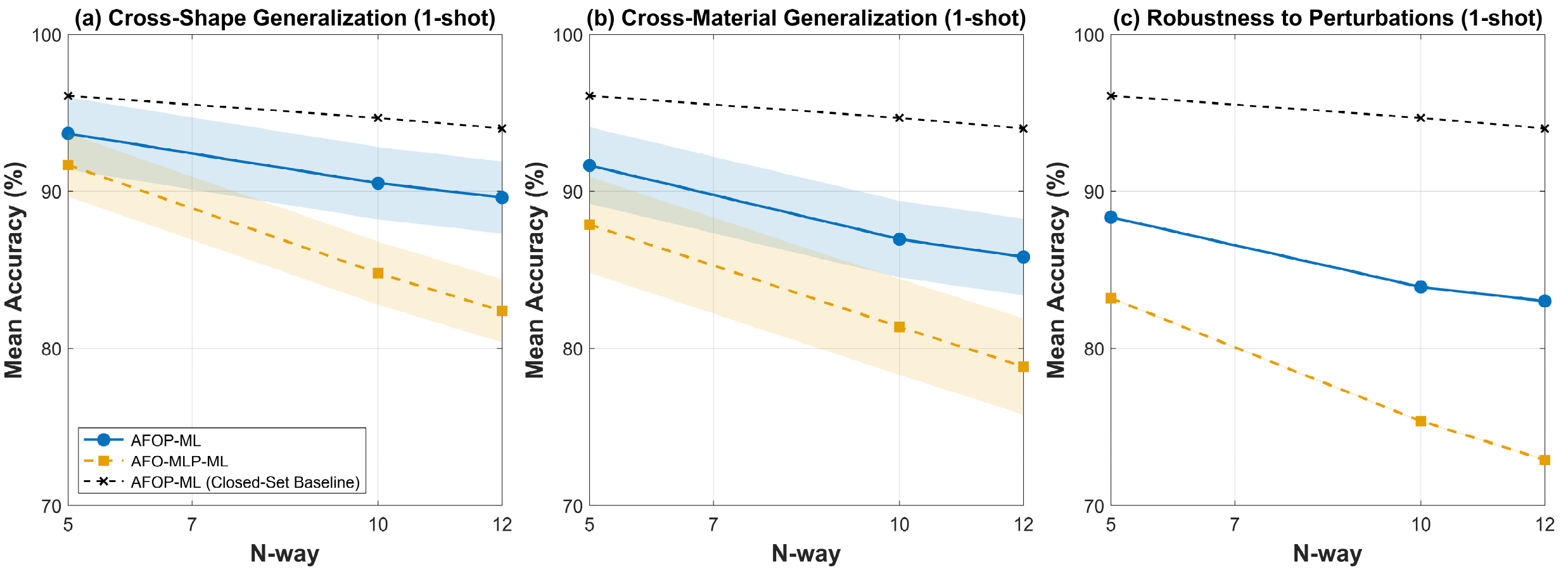}
		\caption{Generalization performance in 1-shot task. (a) Cross‑shape, (b) Cross‑material, and (c) Force–speed perturbations. Shaded bands show the variations within 95\% CI across splits (no split for perturbations).}
		\label{fig:generalization}
	\end{figure*}
	\section{EXPERIMENTAL EVALUATION}
	\label{sec:experiments}
	
	\subsection{Experimental Configuration}
	\label{subsec:config} 
	
	The performance of our AFOP-ML framework will be verified under four different configurations: 1) Closed-Set: 36 classes are split into training and testing samples as per a ratio of 0.5:0.5 per class. Offline training on the training split; Evaluation on the test split with episodic N-way-K-shot tasks, Learning on the support and evaluating on the queries. 2) Cross-Shape: All 12 shape geometries were split into 3 folds. In each fold, 8 shapes are used for offline training and 4 left for testing. \textit{No information about the 4 test shapes was included in training dataset.} 3) Cross-Material: Leave-one-material-out over resin, wood, and aluminum. Only 1 material was used for offline training and other 2 left for testing. \textit{No information about the 2 test materials was included in training dataset.} 4) Force/Speed Perturbation: Offline training at nominal force and speed on the original training set; testing on multiple perturbed force–speed combinations.
	
	\subsection{Evaluation Criteria}
	
	All the results are averaged over 500 episodes with 95\% confidence intervals (CI). Consumption time (or Latency) is reported in offline training and per-episode adaptation, and queries are never used for selection or early stopping. Pretrain time shows the one-time offline cost, and adapt/episode time refers to the adaption time on a 5-way-5-shot episode.
	All results were obtained on an Intel Core i9-12900HX CPU and an NVIDIA RTX 3080 Ti (16 GB) GPU.
	
	In order to validate the effects of our method, the results of two DL method without meta-learning, i.e., CNN and BiLSTM (Bidirectional Long Short-Term Memory), and four other meta-learning approaches are also presented: AFO-MLP-ML, Direct-Prototypical-ML, MAML, and CWT-ResNet-ML. In AFO-MLP-ML, a small 3-layer Multi-Layer Perceptron (MLP) (D$\rightarrow$64$\rightarrow$32$\rightarrow$N with ReLU) is inserted before the prototypical head to nonlinearly project the Top-D embeddings, echoing TapNet’s task-adaptive projection idea \cite{yoon2019tapnet}. Direct-Prot-ML directly feed all the 386 features in feature pool to the prototypical meta-learning network without feature optimization. Model-Agnostic Meta-Learning (MAML) employs CNN to learn features directly and then is fine-tuned with a few gradient steps after pretraining \cite{qiao2025fault}. CWT-ResNet-ML generates the features from a pretrained Resnet after a conversion via CWT \cite{qin2024surface}.
		
	\subsection{Closed-Set Performance: Accuracy-Efficiency}
	\label{subsec:closed_set}
	
	Specific closed-set results are given in Table.~\ref{tab:closed_set_results}. As for 1-shot learning, our proposed AFOP-ML approach achieves $96.1\%$ in 5-way case and maintains $88.7\%$ in the extreme 36-way case, which outperforms the other algorithms. As the number of classes increases $(N=5\to36)$, it exhibits the smallest per-class decline rate $(\approx -0.24~\text{pp/class})$, whereas AFO-MLP-ML and MAML drop substantially faster. Without Meta-learning, the accuracy for CNN and BiLSTM was found as low as 14.14\% and 16.78\% for 36-way recognition. It indicates that AFOP-ML performs excellent and stays the best in all-way recognition when only one example is available for training.
	
	When more labeled examples joined the training set, CWT-ResNet-ML and our AFOP-ML show comparable accuracy while CWT-ResNet-ML obtains slightly higher accuracy in most situations except for 36-way-3-shot. However, CWT-ResNet-ML requires much more time in the pretraining and adaptation stage, respectively 8 minutes and 1428.30 ms. AFO-MLP-ML achieves relatively high accuracy, whereas it requires much more adaptation time. MAML obtains the worst results in 3-shot and 5-shot learning among all meta-learning algorithms. It consumes 20 minutes in pretraining stage and only 72.87 ms in adaptation process. When all the features are directly fed to Prototypical network, i.e., Direct-Prot-ML approach, both the accuracy and computational cost stay at a medium level among all meta-learning methods. In contrast to meta-learning, conventional DL models, e.g., CNN and BiLSTM, show much poorer performance in recognition accuracy and time cost.	
	
	\subsection{Generalization to Unseen Domains (1-Shot)}
	\label{subsec:generalization}
		
	To evaluate the generalization of AFOP-ML, the recognition task is tested under three different scenarios in 1-shot situation (Fig.~\ref{fig:generalization}). The unseen cases mean they don't appear in the meta-training phase and only exist in meta-testing.
		
	\textbf{Cross-Shape Generalization.}  As shown in Fig.~\ref{fig:generalization}(a), the AFOP-ML curve (blue) shows only a minor accuracy drop compared to the closed-set baseline (dashed black line), declining by 2.4 percentage points (pp) at 5-way recognition. It indicates that the learned features effectively capture material-invariant geometric properties and the knowledge learned from existing shapes can be transferred to learn unseen shape geometries rapidly with only 1 example. Furthermore, the performance gap between AFOP-ML and its non-linear counterpart, AFO-MLP-ML, becomes enlarged with task complexity, reaching 7.2 pp in the 12-way scenario, suggesting higher stability of the AFOP-ML framework.
	
	\textbf{Cross-Material Generalization.} As revealed by \cite{kaboli2016re, bauml2019deep}, tactile signals can be highly characterized by object material. Hence, we here test the generalization of our method with different materials. As shown in Fig.~\ref{fig:generalization}(b), the accuracy degradation resulted from material is further aggravated compared with cross-shape case, decreasing by 4.4 pp in 5-way. Nevertheless, it still outperforms the AFO-MLP-ML approach, with the performance gap increased to 7.0 pp in 12-way. Actually, a larger feature dimensionality $D=12$ is learned for this task, among which more features from PVDF are employed. It agrees well with existing works in that PVDF plays a significant role in material discrimination \cite{qin2023perception, qin2017enhanced}.
	
	\textbf{Force/Speed Perturbations.} In practical applications, the applied force and sliding speed often vary, which affects the effectiveness of tactile recognition. As given in Fig.~\ref{fig:generalization}(c), the largest reduction of recognition accuracy can be found in this scenario, i.e., about 7.7 pp in the 5-way task. The performance gap between AFOP-ML and AFO-MLP-ML expands to maximally 10.1 pp in 12-way, confirming that the simpler metric-based head better suited the handling of statistical variations introduced by dynamic physical perturbations.	
	
	\subsection{Interpretability of the Adaptive Framework}
	\label{subsec:interpretability}
	
	\begin{figure}[h]
		\centering
		\includegraphics[width=0.9\columnwidth]{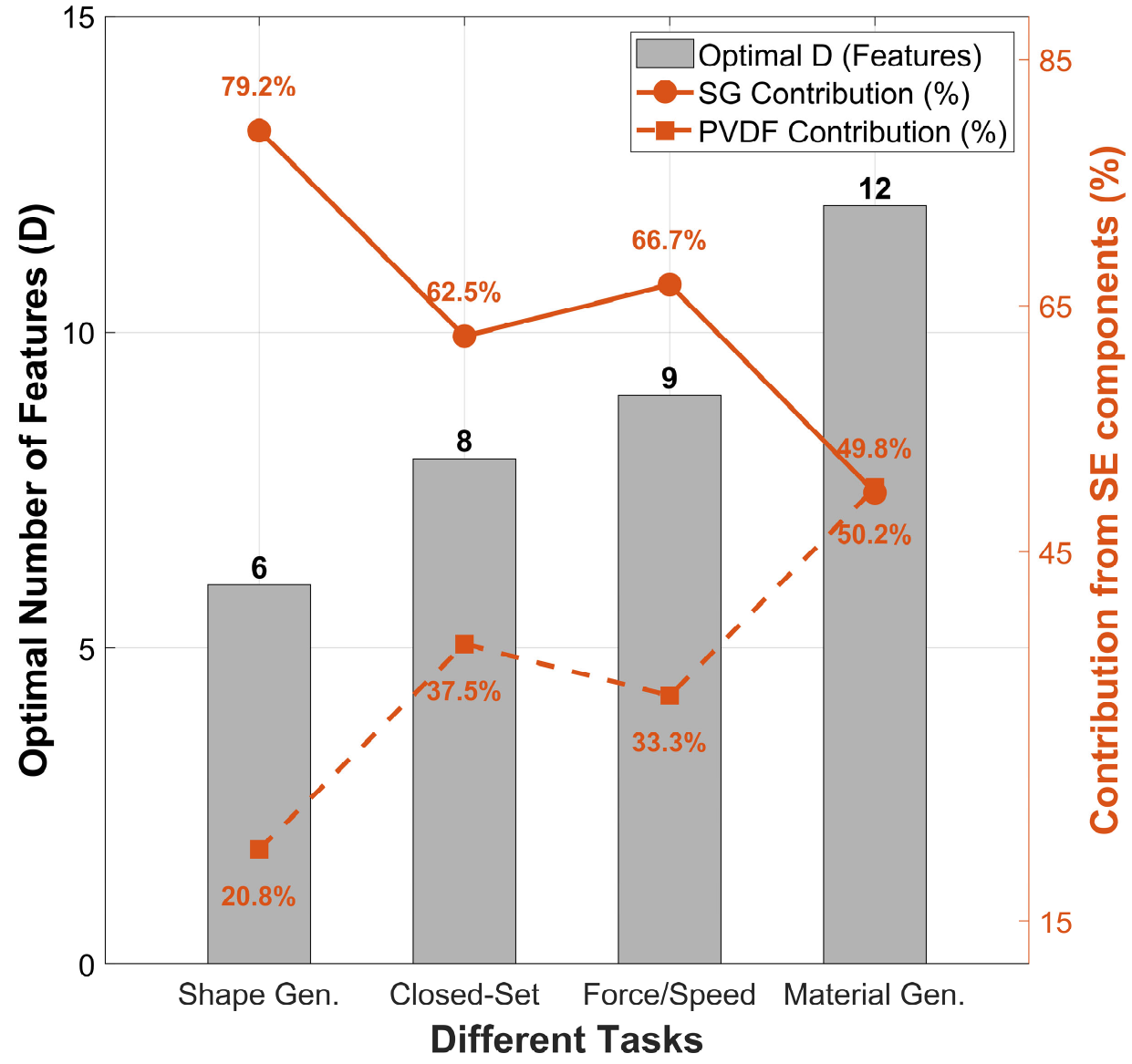}
		\caption{Adaptive determination of feature dimensionality ($D$) and SE contributions in each groups of task. Median number of selected features $D$ (bars, left vertical axis) and the proportion of PVDF-SG-based features among the selected set (line, right vertical axis) across four settings: Shape, Closed-set, Force\&Speed, and Material.}
		\label{fig:interpretability}
	\end{figure}
	
	The key to the framework's robustness lies in its adaptive mechanism, which automatically adjusts both model capacity and sensory focus in response to task demands, as illustrated in Fig.~\ref{fig:interpretability}. The model's capacity, i.e., the optimal feature dimension ($D$), starts from $D \approx 6$ for shape generalization, rises to $D \approx 8$ for the closed-set, and further to $D \approx 9$ and $D \approx 12$ for the more challenging force/speed perturbation and material generalization tasks. It in general reflects that the required number of features varies as per specific recognition task. Meanwhile, the contribution of different SE components also varies with respect to the recognition task. SG plays a much larger role (79.2\%) in cross-shape learning task while PVDF increases from 20.8\% to 50.2\% in Cross-material task. It is also physically intuitive: geometry-driven tasks more rely on low-frequency deformation signals (SG), while material-driven tasks necessitate the inclusion of texture-related vibrational signatures (PVDF).

	\begin{figure}[htbp]
		\centering
		\includegraphics[width=0.9\columnwidth]{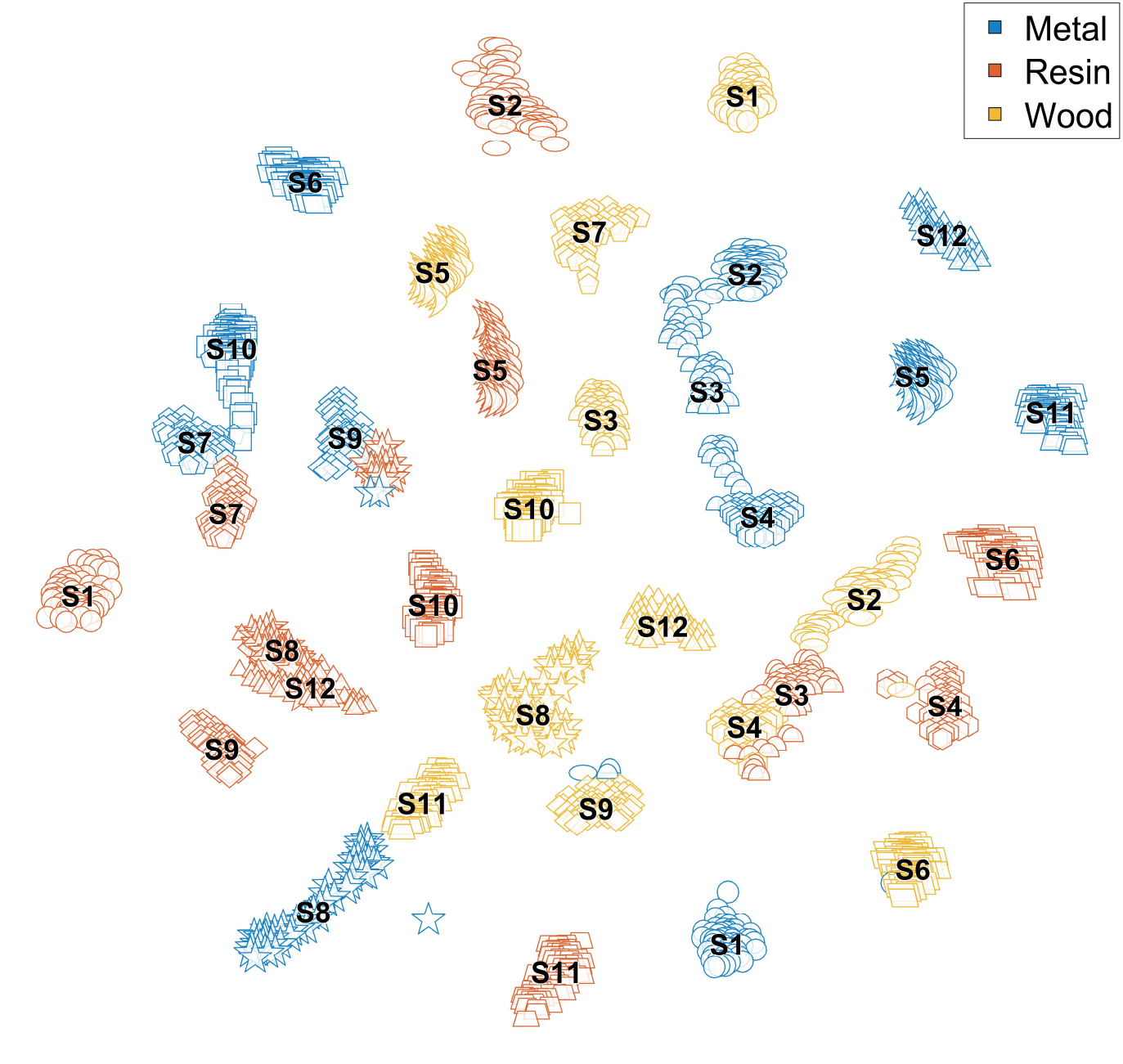}
		\caption{t‑SNE of the closed‑set features ($D=8$, 5-way-5-shot); marker encodes shape (12 class‑specific markers) and color encodes material. 
		Shape clusters are compact and materials are well mixed within each shape cluster. 
		Quantitative indicators: 
		1‑NN=0.982 (nearest‑neighbour shape accuracy), 
		mix‑sil=0.840 (0–1 material‑mixing score; the higher is better), 
		DGI=4.769 (ratio of cross‑material to same‑material neighbors; $>1$ indicates strong mixing).
		t-SNE is used only for qualitative visualization; overlaps in 2D do not necessarily imply misclassification in the original Top-$D$ space, where queries are classified by cosine similarities to class prototypes. }
		\label{fig:tsne}
	\end{figure}
	
	The optimal feature combination determined by AFOP-ML contains 8 features in the 5-way-5-shot closed-set task. To visualize the sample distribution in feature space, t-SNE visualization is applied as shown in Fig.~\ref{fig:tsne}. It can be found that the data forms twelve overarching, well-separated clusters, each corresponding to a unique shape from S1 through S12. The 1-NN indicator (0.982) shows that the learned features have created a space where geometries are highly separable. Simultaneously, high material-mixing scores, validated by a mix-sil of 0.840 and a Distance to Geometric Impostor (DGI) of 4.769, provide quantitative proof that the representation is largely invariant to material. These metrics collectively offer a clear rationale for our generalization results: the high geometric separability directly enables the strong cross-shape transfer (Fig.~\ref{fig:generalization}(a)), while a part of indistinguishable material samples explains why cross-material generalization (Fig.~\ref{fig:generalization}(b)) is inherently more challenging.

	\section{CONCLUSION}
	\label{sec:conclusion}
	
	In this work, we proposed an automatic feature optimization-enabled Prototypical network, i.e., AFOP-ML, for meta-learning based tactile recognition of both shapes and materials with scarce data. In addition to the adaptation of meta-learning model parameters, the optimal feature space was also determined automatically as per specific tasks. Evaluations were implemented in multiple fields, including the closed-set recognition, generalization to unseen domains, and interpretability of the optimal feature space and SE contributions in tactile sensor. While our proposed framework outperformed the other algorithms in 1-shot learning, it just consumed little computational time. Strong generalization was also validated through a series of experiments. Our work not only provides insights for the rapid learning and knowledge transfer between similar domains, but also reveals the significance of sensory components to motivate improved design of tactile sensors. Future work will extend the evaluation to more complex object geometries and broader tactile perception tasks.
	
	\addtolength{\textheight}{-12cm}   
	\bibliographystyle{IEEEtran} 
	\bibliography{reference}

@inproceedings{bauml2019deep,
  title={Deep n-shot transfer learning for tactile material classification with a flexible pressure-sensitive skin},
  author={B{\"a}uml, Berthold and Tulbure, Andreea},
  booktitle={2019 International Conference on Robotics and Automation (ICRA)},
  pages={4262--4268},
  year={2019},
  organization={IEEE}
}

@inproceedings{bottcher2021object,
  title={Object recognition for robotics from tactile time series data utilising different neural network architectures},
  author={Bottcher, Wolfgang and Machado, Pedro and Lama, Nikesh and McGinnity, Thomas M},
  booktitle={2021 International Joint Conference on Neural Networks (IJCNN)},
  pages={1--8},
  year={2021},
  organization={IEEE}
}

@article{narang2021interpreting,
  title={Interpreting and predicting tactile signals for the syntouch biotac},
  author={Narang, Yashraj S and Sundaralingam, Balakumar and Van Wyk, Karl and Mousavian, Arsalan and Fox, Dieter},
  journal={The International Journal of Robotics Research},
  volume={40},
  number={12-14},
  pages={1467--1487},
  year={2021},
  publisher={SAGE Publications Sage UK: London, England}
}

@inproceedings{el2024optimizing,
  title={Optimizing BioTac Simulation for Realistic Tactile Perception},
  author={El Amri, Wadhah Zai and Navarro-Guerrero, Nicol{\'a}s},
  booktitle={2024 International Joint Conference on Neural Networks (IJCNN)},
  pages={1--8},
  year={2024},
  organization={IEEE}
}

@inproceedings{baishya2016robust,
  title={Robust material classification with a tactile skin using deep learning},
  author={Baishya, Shiv S and B{\"a}uml, Berthold},
  booktitle={2016 IEEE/RSJ International Conference on Intelligent Robots and Systems (IROS)},
  pages={8--15},
  year={2016},
  organization={IEEE}
}

@article{wang2022fabric,
  title={Fabric classification using a finger-shaped tactile sensor via robotic sliding},
  author={Wang, Si-ao and Albini, Alessandro and Maiolino, Perla and Mastrogiovanni, Fulvio and Cannata, Giorgio},
  journal={Frontiers in Neurorobotics},
  volume={16},
  pages={808222},
  year={2022},
  publisher={Frontiers Media SA}
}

@article{qin2024surface,
  title={Surface Recognition With a Tactile Finger Based on Automatic Features Transferred From Deep Learning},
  author={Qin, Longhui and Shi, Xiaowei and Yang, Wenhui and Qin, Zhengxu and Yi, Zhengkun and Shen, Huimin},
  journal={IEEE Transactions on Instrumentation and Measurement},
  year={2024},
  publisher={IEEE}
}

@article{shi2023surface,
  title={Surface recognition with a bioinspired tactile fingertip},
  author={Shi, Xiaowei and Wang, Yihua and Qin, Longhui},
  journal={IEEE Sensors Journal},
  volume={23},
  number={16},
  pages={18842--18855},
  year={2023},
  publisher={IEEE}
}

@article{qin2023perception,
  title={Perception of static and dynamic forces with a bio-inspired tactile fingertip},
  author={Qin, Longhui and Shi, Xiaowei and Wang, Yihua and Zhou, Zhitong},
  journal={Journal of Bionic Engineering},
  volume={20},
  number={4},
  pages={1544--1554},
  year={2023},
  publisher={Springer}
}

@article{snell2017prototypical,
  title={Prototypical networks for few-shot learning},
  author={Snell, Jake and Swersky, Kevin and Zemel, Richard},
  journal={Advances in neural information processing systems},
  volume={30},
  year={2017}
}

@inproceedings{finn2017model,
  title={Model-agnostic meta-learning for fast adaptation of deep networks},
  author={Finn, Chelsea and Abbeel, Pieter and Levine, Sergey},
  booktitle={International conference on machine learning},
  pages={1126--1135},
  year={2017},
  organization={PMLR}
}

@inproceedings{li2018slip,
  title={Slip detection with combined tactile and visual information},
  author={Li, Jianhua and Dong, Siyuan and Adelson, Edward},
  booktitle={2018 IEEE International Conference on Robotics and Automation (ICRA)},
  pages={7772--7777},
  year={2018},
  organization={IEEE}
}

@article{nichol2018reptile,
  title={Reptile: a scalable metalearning algorithm},
  author={Nichol, Alex and Schulman, John},
  journal={arXiv preprint arXiv:1803.02999},
  volume={2},
  number={3},
  pages={4},
  year={2018}
}

@article{hospedales2020meta,
	title={Meta-learning in neural networks: A survey},
	author={Hospedales, Timothy and Antoniou, Antreas and Micaelli, Paul and Storkey, Amos},
	journal={IEEE Transactions on Pattern Analysis and Machine Intelligence},
	volume={44},
	number={9},
	pages={5149--5169},
	year={2020},
	publisher={IEEE}
}

@article{vettoruzzo2024advances,
  title={Advances and challenges in meta-learning: A technical review},
  author={Vettoruzzo, Anna and Bouguelia, Mohamed-Rafik and Vanschoren, Joaquin and R{\"o}gnvaldsson, Thorsteinn and Santosh, KC},
  journal={IEEE transactions on pattern analysis and machine intelligence},
  volume={46},
  number={7},
  pages={4763--4779},
  year={2024},
  publisher={IEEE}
}

@inproceedings{kaboli2016re,
  title={Re-using prior tactile experience by robotic hands to discriminate in-hand objects via texture properties},
  author={Kaboli, Mohsen and Walker, Rich and Cheng, Gordon},
  booktitle={2016 IEEE International Conference on Robotics and Automation (ICRA)},
  pages={2242--2247},
  year={2016},
  organization={IEEE}
}

@article{pyo2021recent,
  title={Recent progress in flexible tactile sensors for human-interactive systems: from sensors to advanced applications},
  author={Pyo, Soonjae and Lee, Jaeyong and Bae, Kyubin and Sim, Sangjun and Kim, Jongbaeg},
  journal={Advanced Materials},
  volume={33},
  number={47},
  pages={2005902},
  year={2021},
  publisher={Wiley Online Library}
}

@article{roberge2023stereotac,
  title={StereoTac: A novel visuotactile sensor that combines tactile sensing with 3D vision},
  author={Roberge, Etienne and Fornes, Guillaume and Roberge, Jean-Philippe},
  journal={IEEE Robotics and Automation Letters},
  volume={8},
  number={10},
  pages={6291--6298},
  year={2023},
  publisher={IEEE}
}

@article{zhao2025tactile,
  title   = {Tactile Exploration Enabled Shape Recognition with Multi-Perspective Feature Representation},
  author  = {Zhao, Hongliang and Shi, Xiaowei and Yang, Wenhui and Chen, Huayang and Qin, Longhui},
  journal = {IEEE Sensors Journal},
  year    = {2025},
  volume  = {25},
  number  = {20},
  pages   = {38780--38791},
  doi     = {10.1109/JSEN.2025.3605347},
  publisher = {IEEE}
}

@article{qin2017enhanced,
  title={Enhanced surface roughness discrimination with optimized features from bio-inspired tactile sensor},
  author={Qin, Longhui and Yi, Zhengkun and Zhang, Yilei},
  journal={Sensors and Actuators A: Physical},
  volume={264},
  pages={133--140},
  year={2017},
  publisher={Elsevier}
}

@inproceedings{yoon2019tapnet,
	title={Tapnet: Neural network augmented with task-adaptive projection for few-shot learning},
	author={Yoon, Sung Whan and Seo, Jun and Moon, Jaekyun},
	booktitle={International conference on machine learning},
	pages={7115--7123},
	year={2019},
	organization={PMLR}
}

@article{qiao2025fault,
  title={Fault diagnosis for wind turbine generators based on Model-Agnostic Meta-Learning: A few-shot learning method},
  author={Qiao, Likui and Zhang, Yuxian and Wang, Qisen and Li, Donglin and Peng, Shidong},
  journal={Expert Systems with Applications},
  volume={267},
  pages={126171},
  year={2025},
  publisher={Elsevier}
}
	
\end{document}